\documentclass[10pt,journal,compsoc]{IEEEtran}

\usepackage{multirow}

\usepackage{times}
\usepackage{soul}
\usepackage{url}
\usepackage[utf8]{inputenc}
\usepackage{graphicx}
\usepackage{amsmath}
\usepackage{booktabs}
\usepackage{algorithm, algpseudocode}
\usepackage{epsfig}
\usepackage{graphicx}
\usepackage{amssymb}

\usepackage[T1]{fontenc}    
\usepackage{url}            
\date{} 
\usepackage{booktabs}       
\usepackage{amsfonts}       
\usepackage{nicefrac}       
\usepackage{microtype}      
\usepackage{comment}
\usepackage{array}
\usepackage[outercaption]{sidecap}

\DeclareMathOperator*{\argmin}{arg\,min}
\DeclareMathOperator*{\argmax}{arg\,max}
\DeclareMathOperator{\sign}{sign}

\newcommand{\mnistsize}{32pt}
\newcolumntype{M}[1]{>{\centering\arraybackslash}m{#1}}

\newcommand{\etal}{\mbox{\emph{et al.\ }}}
\newcommand{\ie}{\mbox{\emph{i.e.,\ }}}
\newcommand{\eg}{\mbox{\emph{e.g.,\ }}}

\newlength\savewidth\newcommand\shline{\noalign{\global\savewidth\arrayrulewidth
  \global\arrayrulewidth 1pt}\hline\noalign{\global\arrayrulewidth\savewidth}}

\usepackage{color}
\definecolor{citecolor}{RGB}{119,185,0} 
\usepackage[pagebackref=true,bookmarks=false,colorlinks=true,colorlinks,citecolor=citecolor,filecolor=black,urlcolor=black]{hyperref}

\begin{document}

\title{Query Attack via Opposite-Direction Feature:\\ Towards Robust Image Retrieval}

\author{
  Zhedong Zheng, Liang Zheng, Yi Yang,~\IEEEmembership{Senior Member,~IEEE} and Fei Wu,~\IEEEmembership{Senior Member,~IEEE} \\ 
\IEEEcompsocitemizethanks{
\IEEEcompsocthanksitem Z. Zheng and Y. Yang are with the Australian Artificial Intelligence Institute (AAII), University of Technology Sydney, Australia, 2007. E-mail: \href{mailto:Zhedong.Zheng@student.uts.edu.au}{Zhedong.Zheng@student.uts.edu.au} , \href{mailto:Yi.Yang@uts.edu.au}{Yi.Yang@uts.edu.au}
\IEEEcompsocthanksitem L. Zheng is with the Research School of Computer Science, Australian National University, Australia, 0200. E-mail:\href{mailto:liang.zheng@anu.edu.au}{liang.zheng@anu.edu.au}
\IEEEcompsocthanksitem F. Wu is with the College of Computer Science and Technology, Zhejiang University, China, 310027. E-mail:\href{mailto:wufei@cs.zju.edu.cn}{wufei@cs.zju.edu.cn}
}
}

\markboth{Journal of \LaTeX\ Class Files,~Vol.~14, No.~8, August~2015}%
{Shell \MakeLowercase{\textit{et al.}}: Bare Demo of IEEEtran.cls for IEEE Journals}

\IEEEcompsoctitleabstractindextext{%
\begin{abstract}
Most existing works of adversarial samples focus on attacking image recognition models, while little attention is paid to the image retrieval task. 
In this paper, we identify two inherent challenges in applying prevailing image recognition attack methods to image retrieval. 
First, image retrieval demands discriminative visual features, which is significantly different from the one-hot class prediction in image recognition.
Second, due to the disjoint and potentially unrelated classes between the training and test set in image retrieval, predicting the query category from predefined training classes is not accurate and leads to a sub-optimal adversarial gradient. 
To address these limitations, we propose a new white-box attack approach, Opposite-Direction Feature Attack (ODFA), to generate adversarial queries. Opposite-Direction Feature Attack (ODFA) effectively exploits feature-level adversarial gradients and takes advantage of feature distance in the representation space. To our knowledge, we are among the early attempts to design an attack method specifically for image retrieval. When we deploy an attacked image as the query, the true matches are prone to receive low ranks. We demonstrate through extensive experiments that (1) only crafting adversarial queries is sufficient to fool the state-of-the-art retrieval systems; (2) the proposed attack method, ODFA, leads to a higher attack success rate than classification attack methods, validating the necessity of leveraging characteristics of image retrieval; (3) the adversarial queries generated by our method have good transferability to other retrieval models without accessing their parameters, \ie the black-box setting. 
\end{abstract}
}

\maketitle

\IEEEdisplaynotcompsoctitleabstractindextext

\IEEEpeerreviewmaketitle

\section{Introduction}
\IEEEPARstart{H}UMAN can quickly find similar images from limited candidate images, but fail to scale to millions of images. The image retrieval technique is to help user efficiently and effectively find images of interest from a large scale of digital images, which has been applied to many real-world tasks, such as online shopping~\cite{liu2012street,li2019deep,lin2015deep}, and tourism recommendation~\cite{sigurbjornsson2008flickr,wang2017effective,chen2016deep}. Recent advances in this field are due to two factors, \ie the availability of large-scale datasets and the discriminative features extracted from deeply-learned models. The state-of-the-art methods even achieve over 90\% Recall@1 accuracy and surpass the human-level performance~\cite{zhang2017alignedreid,radenovic2018fine,shen2019multi,liu2017provid,li2019deep}. However, one problem remains whether these deeply-learned models are accurate and robust enough to various images in the realistic scenario. The robustness evaluation of the image retrieval system has not been well studied. Inspired by the strong ability of human towards small variants, we investigate the adversarial examples to evaluate the retrieval system via adding small, human-imperceptible noise to the original image. The crafted adversarial examples are to imitate the extreme case in the real applications and cheat the retrieval model of predicting a totally different ranking result. In this way, the adversarial attack serves as an indicator to assess the robustness of the target model and understand the weakness of current image retrieval methods.

\begin{figure*}[t]
\begin{center}
   \includegraphics[width=1\linewidth]{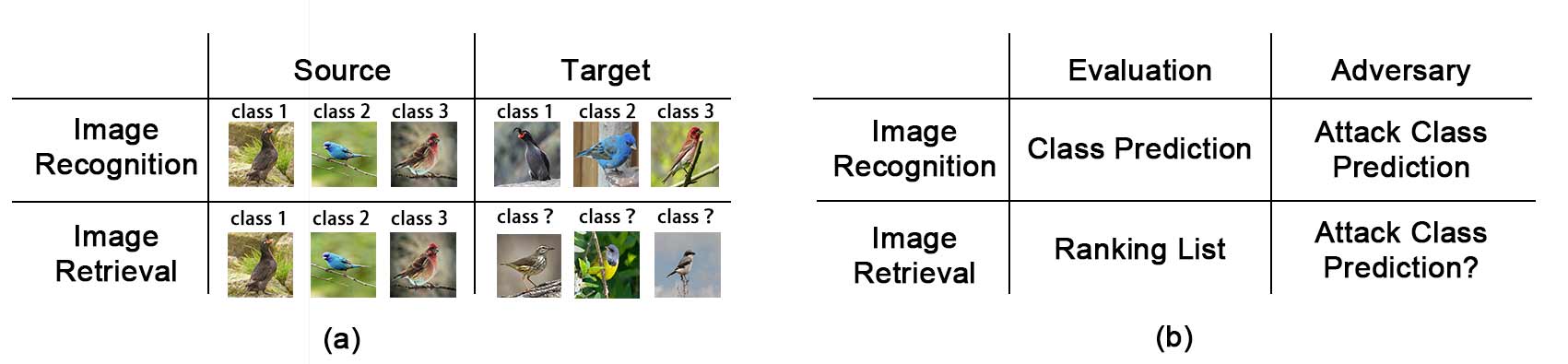}
\end{center}
   \caption{Comparison between image recognition and image retrieval. (a) Problem definition. Image recognition usually recognizes the same classes in the source set and target set. The target set in image retrieval has very few or even no overlapping classes with the source set. (b) The existing classification attack methods employ the classification adversary, which is consistent with its target testing procedure but is inconsistent with the testing procedure in image retrieval. \textbf{In this work, we directly attack the visual feature and explore feature-level adversarial gradients, yielding significant attack success rate on the image retrieval task.} 
   }
\label{fig:problem}
\end{figure*}

For attacking image retrieval models, there is no well-studied method. Most existing methods for generating adversarial examples focus on the classification setting, in which the source and target sets share exactly the same classes~\cite{szegedy2013intriguing,goodfellow2014explaining,kurakin2016adversarial,moosavi2016deepfool,eykholt2018robust,xiao2018generating,xiao2018spatially}. However, in image retrieval, the target set has limited overlap or even no overlap classes with the source set (see Figure~\ref{fig:problem}~(a)) ~\cite{Fu2015Transductive,lin2018unsupervised,sun2019learning,qian2019leader,zhong2020learning,wang2017survey}. 
Instead of predicting the class of the input, image retrieval is to
compute the similarity between the query and database images, and find relevant images according to the similarity score. In this context, we consider the task of generating adversarial examples from the feature space to fool the retrieval system. 

There are two main challenges when deploying existing classification attack methods in the retrieval scenario. First, classification attack methods target class predictions to generate adversarial examples, but this strategy is inconsistent with the testing procedure of image retrieval (see Figure~\ref{fig:problem}~(b)). 
In image retrieval, we aim to retrieve relevant images from the database images by matching visual features. Therefore, attacking the class prediction does not directly affect the retrieval task, which relies on the intermediate deep features. 
Second, classification methods attack the class prediction, which usually does not contain the query class in image retrieval. Given a query image of an unseen class, traditional attack methods lead to the inferior adversarial gradient, which compromises the effectiveness of the attack. 

To address the potential problems of classification attack approaches, this work focuses on generating adversarial examples tailored for image retrieval. 
Image retrieval target catching the true matches in the top-K of the ranking list.
To achieve a successful adversarial attack on the retrieval system, all true matches should be ranked as low as possible in the ranking list. 
In this setting, an alternative solution to attacking the query image is to attack the database (candidate image pool). However, the database is sometimes inaccessible and contains thousands of images. Attacking a large number of database images is prohibitive in terms of time cost. In this paper, therefore, we focus on crafting adversarial query images. Without knowledge of the database, we report that adversarial queries alone are sufficient to fool the retrieval system and the cost of generating an adversarial query is relatively low. 

We propose a white-box attack method for adversarial example generation in the retrieval context, named Opposite-Direction Feature Attack (ODFA). ODFA exploits the gradient in the feature space, and is based on the target testing procedure, \emph{i.e.,} similarity computation between the query and database images using their respective features. 
Our key idea underpinning the attack is to explicitly push the feature of the adversarial example away from its original representation. In particular, we first define the \emph{opposite-direction feature}, which, as its name implies, faces the opposite direction to the feature of the original query. During the adversarial attack, we then force the query feature to move towards the \emph{opposite-direction feature}. Due to the revised direction of the feature vector of the adversarial query, the similarity between the database true matches and the adversarial query will decrease. The retrieval model is therefore prone to treating all the true matches as outliers when an adversarial query is used. 
In the experiment, we show that the proposed ODFA method leads to a large drop in ranking accuracy on five image retrieval datasets, \emph{i.e.,} Food-256~\cite{kawano14c}, CUB-200-2011~\cite{WahCUB_200_2011}, Market-1501~\cite{zheng2015scalable}, Oxford5k~\cite{philbin2007object} and Paris6k~\cite{philbin2008lost}. Under various levels of image perturbation, ODFA outperforms the conventional attack methods, \ie fast-gradient sign method~\cite{goodfellow2014explaining}, basic iterative method~\cite{kurakin2016adversarial} and iterative least-likely class method~\cite{kurakin2016adversarial}. Moreover, when ODFA is adopted on image recognition systems such as ResNet~\cite{he2016deep} on Cifar-10~\cite{Krizhevsky2009Learning}, the effect of its attack does not show clear superiority over this same set of methods~\cite{goodfellow2014explaining,kurakin2016adversarial}. This indicates that the specificity of our method on image retrieval problems.
Additionally, we observe that ODFA has good transferability in the black-box setting; that is, the adversarial queries crafted for one white-box model remain adversarial for another black-box model in the retrieval scenario.
In summary, our contributions are itemized below.
\begin{itemize}
  \item To our knowledge, we are among the early attempts toward attacking image retrieval. While deeply-learned retrieval systems have achieved impressive retrieval performance, they still suffer from the lack of robustness against trivial noise. We find that crafting adversarial queries alone is sufficient to fool the state-of-the-art retrieval systems.
  \item We propose a new adversarial attack method, Opposite-Direction Feature Attack (ODFA), tailored for attacking image retrieval. Specifically, our method works on the feature level (feature matters most for retrieval) instead of the classifier level (classifier matters most for classification). We can thus effectively and efficiently generate the adversarial queries. In addition, we extend ODFA to ODFA-MS for multi-scale inputs.
  \item We conduct experiments on five image retrieval datasets. The experimental results show that the adversarial queries with human-imperceptible noise can successfully cause large performance drops on not only the baseline victim model but also prevailing retrieval approaches. 
  \item Without the knowledge of the target model, the proposed adversarial queries have good transferability and could be applied to the black-box setting.
\end{itemize}

\section{Related Work}
\textbf{Image retrieval.}
Recently the progress in image retrieval has been due to two factors: the availability of large-scale datasets and the learned representation using the deep neural network. Some early works directly apply the off-the-shelf Convolutional Neural Network (CNN) pretrained on the large-scale dataset like ImageNet~\cite{russakovsky2015imagenet} to extract visual representations and sort the candidate images according to the feature similarity~\cite{yue2015exploiting,tolias2015particular,li2019deep,deng2019saliency,yang2018shared,jin2018deep,lin2018unsupervised,yang2017supervised}. The bias between the source set ImageNet and the target datasets, \eg Oxford5k, compromises the ranking performance. Most works, therefore, collect the related large-scale datasets like Landmarks dataset~\cite{babenko2014neural}, in which the data shares similar distribution with the target set, and fine-tune the CNN-based model. The visual representation is tuned in an end-to-end manner and shows a stronger discriminative ability~\cite{radenovic2016cnn,tolias2015particular,yang2018person}. 
The demon is usually in the details, and another line of works is to explore the detailed information~\cite{yu2017devil,liu2017provid,bai2017scalable,wang2020deep}. Some researchers design the objectives to force the CNN model to learn discriminative features and better similarity metric. For instance, triplet loss with hard sampling policy is widely-applied~\cite{ristani2018features,Song2016Deep,zheng2018discriminatively,yu2018hard}. Other works pay attentions to the local patterns, and explicitly involve the image parts into training~\cite{suh2018part,sun2019learning,bai2020deep}. For instance, Radenovi{\'c} \etal~\cite{radenovic2018fine} propose a trainable pooling layer to deal with the scale problem, and arrive the state-of-the-art performance. Zhang \etal~\cite{zhang2017alignedreid} propose a dynamic part-matching method and achieve the result surpassing the human-level performance.

Despite the impressive performance, no prior works have explicitly explored the robustness of the retrieval system. One related practice is to simply collect more distractor images and add them to the testing database, such as 100,000 images, to validate the system robustness~\cite{philbin2007object,philbin2008lost,zheng2015scalable,guo2018learning}. This line of the practice considers to increase the complexity of the database and is orthogonal to our method focusing on the query variants. Compared with our method, increasing the database is limited in two aspects. 1) Most collected images are not ``strong negatives'', which could cheat the retrieval system of changing the ranking list. The true matches with high similarity score are still in the top-K of the ranking results. 
2) Increasing the database also leads to larger time cost. When testing, one needs to extract the feature for a large number of candidate images. 
In contrast, the proposed adversarial queries do not increase the evaluation time cost and efficiently affect the final ranking result.

\noindent\textbf{Adversarial attack.} Adversarial Attack is to craft the sample from the real data to fool the learned model and helps to evaluate the robustness of the target models~\cite{chakraborty2018adversarial,zhang2018face,li2019universal}. Szegedy \emph{et al.} first show that the adversarial images, while looking pretty much the same as the original ones, can mislead the CNN model to classify them into a specific class~\cite{szegedy2013intriguing}. It raises the security problem of the current state-of-the-art models~\cite{sharif2016accessorize,eykholt2018robust} and  also provides us more insights into the CNN mechanism~\cite{goodfellow2014explaining}.
The adversarial attack literature can be broadly divided into two classes, \emph{i.e.,} gradient-based attack and score-based attack.
Given an input image, gradient-based methods need to know the gradient of the applied model. One of the earliest works is the fast-gradient sign method~\cite{goodfellow2014explaining}, which generates adversarial examples in one step. Some works extend~\cite{goodfellow2014explaining} to iteratively updating the adversarial images with small step sizes, \ie basic iterative method~\cite{kurakin2016adversarial}, deep fool~\cite{moosavi2016deepfool} and momentum iterative method~\cite{Dong2017Boosting}. 
Compared with the fast-gradient sign method, the perturbation generated by iterative methods is more smooth, and the adversarial samples are, therefore, more imperceptible to the human.
On the other hand, another line of methods relies on searching the input space. Jacobian-based saliency map attack greedily modifies the input instance~\cite{Papernot2016The}. In~\cite{Narodytska2016Simple}, Narodytska \emph{et al.} further shows that single pixel perturbation, which is out of the valid image range, can successfully lead to misclassification on small-scale images. They also extend the method to large-scale images by local greedy searching. Except for pixel modification, sample generation via spatial transform also can result in the adversarial examples~\cite{xiao2018spatially}. 
The closest inspiring work is the iterative least-likely class method~\cite{kurakin2016adversarial}, which makes the classification model output interesting mistakes, \emph{e.g.,} classifying an image of the class \emph{vehicle} into the class \emph{cat}. They achieve this effect by constraining to increase the predicted probability of the least-likely class. This work adopts a similar spirit. In order to fool the retrieval model into assigning the true matches with possibly low ranks, we constrain to increase the similarity of the query feature vector with a vector of an opposite direction in the feature space. Here we emphasize that our work is different from~\cite{kurakin2016adversarial} in two aspects. First, Kurakin \emph{et al.}~\cite{kurakin2016adversarial} focus on image recognition and rely on class predictions to obtain the least-likely class. In the retrieval setting, the classification model faces images from unseen classes. The inaccurate class prediction compromise the iterative least-likely class method. In this respect, the proposed method directly works on the intermediate feature level and alleviates this problem. Second, Kurakin \emph{et al.}~\cite{kurakin2016adversarial} increase the probability of the least-likely class but do not decrease the probability of the most-likely class. So the true match images / classes are still in the top-K prediction. In comparison, our method explicitly constrains to decrease the similarity of the adversarial image and its original image in the feature space, so that the similarity between the adversarial image and original true-matches also drops. The model is prone to ranking all the true-matches out of the top-K.

\section{Notation}
\subsection{Problem Setup} 
We denote the original query image and its adversarial example as $X$ and $X'$, respectively. Given a query image $X$, the image retrieval model extracts the visual feature $f_X=F(X)$, and then ranks the database images according to the similarity score in the feature space. $F(\cdot)$ denotes a nonlinear mapping function. In this work, we study the widely-used image retrieval models, \ie convolutional neural network (CNN) as the mapping function $F$. For two images $X_m$ and $X_n$, the similarity score in the feature space can be formulated as the cosine similarity: $D(X_m,X_n)=\frac{f_{X_m}}{||f_{X_m}||_2} \times \frac{f_{X_n}}{||f_{X_n}||_2}$, where $||\cdot||_2$ is the L2-norm. 
A high $D$ score indicates that the two images are very similar.
To successfully attack the retrieval result, we intend to find the adversarial query $X'$ to lower the similarity score between the query and all true matches. 
In the mean time, we demand that the difference between the adversarial query and the original query could be as small as possible, which ensures that the adversarial perturbation is imperceptible to the human. In particular, we follow the practice in~\cite{kurakin2016adversarial} to keep pixel difference within a valid value range. We clip the pixels whose values fall out of the valid range, and remove the distortions which are larger than the hyper-parameter perturbation rate $\epsilon$:
$\mbox{Clip}_{X, \epsilon}\{X'\} = \min \{ 255, X+\epsilon, \max \{ 0,X-\epsilon, X'\} \}$. It ensures that $X'$ with $\|X' - X\|_{\infty} \leq \epsilon$.
Since a large $\epsilon$ will make the perturbation perceptible to the human, we set the $\epsilon \le 16$ in this work. 

\subsection{Victim Model}
We call the model to be attacked as the victim model and adopt the white-box assumption as the conventional gradient-based methods~\cite{szegedy2013intriguing,goodfellow2014explaining,kurakin2016adversarial,Dong2017Boosting}; that is, the parameters of the victim model are accessible. Under the assumption, we can obtain the gradient to the inputs. 
To verify the effectiveness of the proposed method, we mainly adopt two kinds of widely-used retrieval models as victim model, \ie \emph{Classification-based retrieval model} and \emph{Ranking-based retrieval model}. The main difference between these two models is whether we can access the category prediction.

\noindent\textbf{Classification-based retrieval model.} Classification-based retrieval model exploits category recognition as the proxy task to learn the projection function from data to the feature space. Following the common practice in~\cite{babenko2014neural,wei2017cross,qian2017multi,li2018learning}, we train the CNN-based model mapping the training data to the class label, and adopt the feature before the final classification layer as the retrieval feature $f$. To compare with the traditional classification attack methods, which rely on the class prediction, we preserve the final classification prediction $p$. The final linear classifier can be formulated as: $p = Wf+b$, which maps the feature $f$ to the class probability $p$. $W$ and $b$ are learned weights of the classification layer.

\noindent\textbf{Ranking-based retrieval model.}  Ranking-based retrieval model is trained with the distance-based objectives, \eg ranking loss~\cite{Song2016Deep, hermans2017defense,zhang2019part} and contrastive loss~\cite{radenovic2018fine}, which pull examples with different class labels apart from each other and push examples from the same classes closer to each other. This line of model does not involve the category prediction part. The conventional classification attack method, therefore, could not work on this kind of models, when our attack method is still feasible. 

To compare with the traditional attacking methods and illustrate our intuition, we assume that the category prediction is available when attacking. We adopt the \emph{Classification-based retrieval model} as the victim model in the following Section~\ref{sec:method}. However, we note that the proposed method does not depend on the category prediction, and also could work on the \emph{Ranking-based retrieval model}. In  Experiment, we show the result of the proposed model attacking both kinds of retrieval models. 

\section{Method} \label{sec:method}
In this section, we first extend the traditional adversarial methods to the retrieval scenario and discuss the limitation of these methods. We next introduce the proposed Opposite-Direction Feature Attack (ODFA) method, which addresses the weakness of the traditional methods. Furthermore, we extend ODFA to ODFA-MS, attacking the common evaluation trick, \ie the feature fusion of multi-scale inputs.

\subsection{Adoption of Classification Attack in Image Retrieval}
\label{sec:old}
Previous works in the adversarial example generation are designed for image recognition and aim to attack the class prediction~\cite{goodfellow2014explaining,kurakin2016adversarial}. When the prediction is changed, the intermediate activation in the network is also implicitly impacted. Although these methods are not designed for the retrieval problem, they still work for the retrieval scenario with a minor modification. We assume that we could acquire the label prediction of the victim model, and extend these existing classification attack methods to generate the adversarial queries. 

Specifically, for the fast-gradient sign method~\cite{goodfellow2014explaining} and basic iterative method~\cite{kurakin2016adversarial}, we deploy the label predicted by the victim model as the pseudo label
$ y_{max} = \argmax_{y} \bigl\{p(y|X) \bigr\}$. To fool the model, the objective is to decrease the probability $p(y_{max})$ that the adversarial query $X'$ is classified into the pseudo class. The objective is written as,
\begin{equation}
\argmin_{X'} J(X') = log(p(y_{max}|X')) .
\end{equation}
For the iterative least-likely class method~\cite{kurakin2016adversarial}, we calculate the least-likely class $ y_{min} = \argmin_{y} \bigl\{p(y|X)\bigr\}.$ The attack objective is to increase the probability $p(y_{min})$ so that the input is classified as the least-likely class. The objective is,
\begin{equation}
\argmax_{X'} J(X') = log(p(y_{min}|X')) .
\end{equation}
When generating adversarial samples, the weight of the victim model is fixed and we only update the input. 
For the fast-gradient sign method, $X' = X + \epsilon \sign(\nabla J(X))$. 
For the iterative methods, \emph{i.e.,} basic iterative method and iterative least-likely class method, 
we initialize $X'$ with $X$: $X'_{0} = X$, and then update the adversarial samples $T$ times: $X'_{t+1} = X'_{t} + \alpha \sign(\nabla J(X'_{t}))$, where $\alpha$ is a relatively small hyper-parameter. Following the practice~\cite{kurakin2016adversarial}, we set $\alpha = 1 $ and the number of the iterations $T = min(\epsilon+4,1.25 \times \epsilon)$. The clip function $Clip_{X,\epsilon}\{X'\}$ is also added in every iteration to keep pixels of the adversarial query in the valid range.

\noindent\textbf{Discussion.} \emph{Why the conventional classification attack methods can work for retrieval?}
The retrieval system learns a semantic projection function, mapping input images to the feature space, which is highly relevant to the semantics category. Although classification attack methods do not target changing the representation, they make changes to the category prediction $p$ of the query. In particular, according to the prediction function $p=Wf+b$ (note that $W$ and $b$ are fixed), the intermediate feature $f$ is also implicitly affected when optimizing the adversarial objective. Due to the changes to the feature, the semantic similarity between the adversarial example and the original image implicitly decreases. The traditional attack methods, therefore, could generate effective adversarial samples for the retrieval system.

\emph{What are the disadvantages of the classification attack for retrieval?} There are two main disadvantages. First, the source set  and the query usually do not contain the same set of classes. The predefined training classes in the source set cannot well represent the semantics of the unseen query. So the predicted most-likely label may not really be the most-likely one, and the predicted least-likely label may not really be the least-likely one, either. Second, the above-mentioned three classification attack methods~\cite{kurakin2016adversarial,goodfellow2014explaining} work on the prediction score and do not explicitly change the similarity in the feature space. The classification attack, therefore, usually obtains a sub-optimal gradient and is limited in their adversarial performance on the retrieval system.

\begin{figure*}[t]
\begin{center}
   \includegraphics[width=0.95\linewidth]{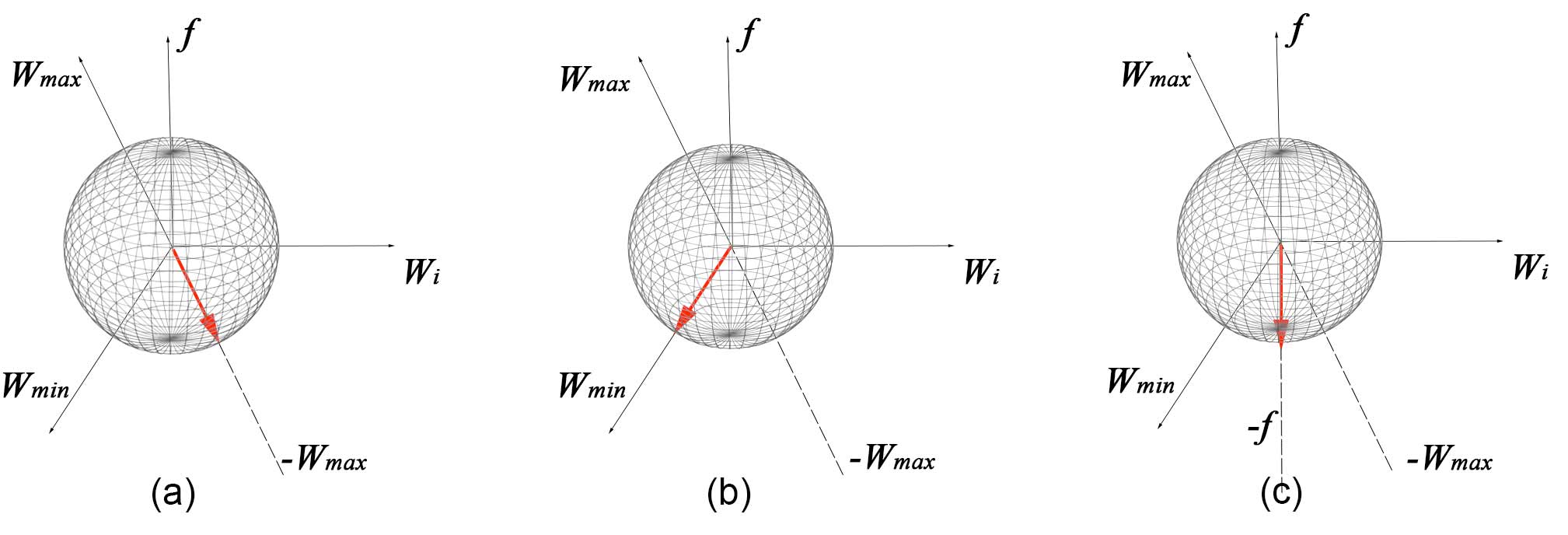}
\end{center}
   \caption{Geometric interpretation of (a) the fast-gradient sign method \protect\cite{goodfellow2014explaining} and the basic iterative method \protect\cite{kurakin2016adversarial}, (b) the iterative least-likely class method \protect\cite{kurakin2016adversarial}, and (c) the proposed ODFA. The red arrows represent the direction of the gradient on the original feature $f$. $W_{max}$ denotes the weight of most-likely class $y_{max}$ and $W_{min}$ denotes the weight of the least-likely class $y_{min}$. The proposed method does not rely on the class prediction scores and deploys a straightforward opposite gradient direction $-f$ in attacking the retrieval features.}
\label{Geometric}
\end{figure*}

\begin{algorithm}[t]
\small
\caption{Opposite-Direction Feature Attack (ODFA)}
\label{alg:ODFA}
\begin{algorithmic}[1]
\Require The victim model $F$; a real query image $X$;
\Require The perturbation rate $\epsilon$.
\Ensure
An adversarial example $X'$ with $\|X' - X\|_{\infty} \leq \epsilon$.
\State $X_0' = X$;
\State $T = min(\epsilon+4,1.25 \times \epsilon)$;
\For {$t = 0$ to $T-1$}
\State Input $X_t'$ to $F$, extract feature $f$, calculate the objective $J(X_t')$;
\vspace{-2ex}
\begin{equation}
J(X'_t) = ( \frac{f_{X'_t}}{||f_{X'_t}||_2} + \frac{f_X}{||f_X||_2})^2.
\label{eq:loss}
\end{equation}
\vspace{-2ex}
\State Update $X_{t+1}'$ by applying the sign gradient as 
\vspace{-1.5ex}
\begin{equation}
\label{eq:update}
X'_{t+1} = X'_t + \alpha \sign(\nabla J(X'_{t})).
\end{equation}
\State Keep pixels of the adversarial query in the valid range
\begin{equation}
X'_{t+1} = \mbox{Clip}_{X, \epsilon}\{X'_{t+1}\}. 
\label{eq:clip}
\end{equation}
\vspace{-4ex}
\EndFor \\
\Return $X' = X_T'$.
\end{algorithmic}
\end{algorithm}

\subsection{Opposite-Direction Feature Attack}
\label{sec:new}
To overcome the above disadvantages of the classification attack methods, we propose a new attack method named Opposite-Direction Feature Attack (ODFA), which directly works on the intermediate feature without the prerequisite to the category predictions. 
Specifically, given a query image $X$, the retrieval model extracts the original feature $f_{X}$. We argue that the similarity score $D(X, X_{gt})$ between query $X$ and its true match $X_{gt}$ is relatively high in a well-learned retrieval system. To attack the retrieval model, our target is to minimize the similarity score $D(X',X_{gt})$ between the adversarial query $X'$ and its true match image $X_{gt}$. To achieve this goal, we define the loss objective as,
\begin{equation}
\argmin_{X'} J(X') = ( \frac{f_{X'}}{||f_{X'}||_2} + \frac{f_X}{||f_X||_2})^2.  \label{Eq:our}
\end{equation}
This loss function aims to push the feature $f_{X'}$ of the adversarial image to the opposite side of the original query feature $f_{X}$. We name $-f_{X}$ as the \emph{opposite-direction feature}. When the objective $J(X') \to 0$ , $\frac{f_{X'}}{||f_{X'}||_2}$ will be close to $-\frac{f_X}{||f_X||_2}$, $D(X,X')\to-1$. The similarity score between the adversarial query and the true match images is,
\begin{equation}
\begin{aligned} 
D(X',X_{gt}) &= \frac{f_{X'}}{||f_{X'}||_2} \times \frac{f_{X_{gt}}}{||f_{X_{gt}}||_2} \to \\
- \frac{f_{X}}{||f_{X}||_2}\times \frac{f_{X_{gt}}}{||f_{X_{gt}}||_2} &= -D(X,X_{gt}).
\end{aligned} 
\end{equation}
Since $D(X,X_{gt})$ is usually high in the original retrieval model, we can deduce that the similarity score $D(X',X_{gt})$ is low. To generate an adversarial query $X'$, we adopt an iterative method to update $X'$: $X'_{0} = X, X'_{t+1} = X_{t} + \alpha \sign(\nabla J(X'_{t}))$. The clip function is also added to keep pixels in the adversarial sample within the valid range. The overall process of crafting the adversarial query is present in Algorithm~\ref{alg:ODFA}. 

\noindent\textbf{Discussion.} We provide a 2D geometric interpretation to illustrate the difference of the gradient direction between the proposed method and traditional attack methods (see Figure~\ref{Geometric}). The classification attacks use the class prediction $p = Wf+b$, where $W$ is the learned weight and $b$ is the bias term. The weight $W=\{W_1,W_2,\dots,W_K\}$ contains $K$ weights for the $K$ classes in the source set. We use $W_{max}$ to denote the weight of most-likely class $y_{max}$ and $W_{min}$ to denote the weight of the least-likely class $y_{min}$. 
For the fast-gradient sign method and the basic iterative method, the gradient on feature $f$ equals to,
\begin{equation}
\frac{\partial J(X')}{\partial f_{X'}} = - W_{max}\times \frac{\partial J(X')}{\partial p(y_{max})}.
\end{equation}
Note that $\frac{\partial J(X')}{\partial p(y_{max})}$ is a positive constant. So the direction of the gradient is the direction of $-W_{max}$ (see Figure~\ref{Geometric}~(a)). For the iterative least-likely class method, the gradient equals to,
\begin{equation}
\frac{\partial J(X')}{\partial f_{X'}} = W_{min}\times \frac{\partial J(X')}{\partial p(y_{min})}.
\end{equation}
The gradient has the same direction with $W_{min}$ (see Figure~\ref{Geometric}~(b)). For the unseen images of new classes, \emph{i.e.,} query images, $-W_{max}$ and $W_{min}$ are not accurate to describe the adversary of the original query, so the adversarial attack effect is limited. In this paper, instead of using class predictions, we directly attack the representation in the feature space. According to the Eq.~\ref{Eq:our}, the adversarial gradient of the proposed method is written as,
\begin{equation}
\frac{\partial J(X')}{\partial f_{X'}} = -2\times(\frac{f_{X'}}{||f_{X'}||_2}+\frac{f_{X}}{||f_{X}||_2}),
\end{equation}
where $f_{X}$ is the feature of the original query image. In Figure~\ref{Geometric} (c), we draw the gradient direction of the first iteration. In the first iteration, $f_{X'_0} = f_{X}$, $\frac{\partial J(X'_0)}{\partial f_{X'_0}} = -4\frac{f_{X}}{||f_{X}||_2}$. Our method leads the feature to the opposite direction of the original feature, so the similarity of true matches drops more quickly. The observation in the experiment, as shown in Figure~\ref{market} and Figure~\ref{fig:CUB}, also verifies that the proposed method is more efficient than the conventional methods.

\begin{algorithm}[t]
\small
\caption{Opposite-Direction Feature Attack with Multiple-Scale Inputs (ODFA-MS)}
\label{alg:ODFA-MS}
\begin{algorithmic}[1]
\Require The victim model $F$; a real query image $X$;
\Require The multiple-scale factors $S$; The perturbation rate $\epsilon$.
\Ensure
An adversarial example $X'$ with $\|X' - X\|_{\infty} \leq \epsilon$.
\State $X_0' = X$;
\State $T = min(\epsilon+4,1.25 \times \epsilon)$;
\For {$t = 0$ to $T-1$}
\State Resize the $X_t'$ to ${X_t^s}'$ for $s$ in $S$
\State Input ${X_t^s}'$ to $F$, extract features of different scales $f_{{X_t^s}'} = F({X_t^s}').$
\State Calculate the objectives $J({X_t^s}')$ as Eq.~\ref{eq:loss} and gradients of different-scale inputs;
\State Resize the gradients to the original scale and average the gradients
\begin{equation}
\nabla J(X'_{t}) = \frac{1}{n_s} \sum {\nabla \tilde{J}({X_t^s}')}.
\end{equation}
\State Update $X_{t+1}'$ by applying the sign gradient as Eq.~\ref{eq:update}
\State Keep pixels of the adversarial query in the valid range as Eq.~\ref{eq:clip}
\EndFor \\
\Return $X' = X_T'$.
\end{algorithmic}
\end{algorithm}

\subsection{Opposite-Direction Feature Attack with Multiple-Scale Inputs }
Fusing the features of multiple-scale inputs is a common practice in many image retrieval systems, such as landmark instance retrieval~\cite{radenovic2016cnn,radenovic2018fine}. In particular, when testing, the input image is resized with multiple scale factors $S=\{s_1, s_2, ..., s_{n_s}\}$, and then the model extracts the feature from inputs of different scales. The mean of the normalized features is used as the final retrieval representation. 
Since the final representation fuses the feature of the multi-scale inputs, the retrieval system is more robust in terms of the scale variants. In the experiment, we observe that only calculating the adversarial gradient upon the input of the original scale is less effective to fool the image retrieval system. It is due to that the designed imperceptible perturbation is deprecated when resizing images.

To successfully attack the multiple-scale inputs, we further extend the proposed ODFA to ODFA-MS. The whole pipeline is summarized in Algorithm~\ref{alg:ODFA-MS}. We first view the inputs of different scales as independent inputs $X_t^s$. Similar to the single scale setting, we calculate the adversarial gradient based on each scale $\nabla J({X_t^s}')$. To generate the adversarial gradient towards the original input, we resize all gradients to the original scale $\nabla \tilde{J}({X_t^s}')$ and average the multi-scale adversarial gradients as follow, 
\begin{equation}
\nabla J(X'_{t}) = \frac{1}{n_s} \sum {\nabla \tilde{J}({X_t^s}')},
\end{equation}
in which $n_s$ is the number of scale factors. Similar to ODFA, we add the sign gradient to the original input and iteratively update the input to obtain the adversarial samples. Since we explicitly consider multi-scale adversarial gradients, the ODFA-MS significantly outperform the regular ODFA in terms of multiple-scale evaluation. More details can be found in Section~\ref{subsec:ranking}.

\section{Experiment}

\subsection{Datasets and Settings}
We evaluate the attack performance on five image retrieval datasets, \ie Food-256, CUB-200-2011, Market-1501, Oxford5k, Paris6k, and one image recognition dataset, \ie Cifar-10.

\noindent\textbf{Food-256 }is a large-scale food retrieval dataset~\cite{kawano14c}.
The author collects images of 256-kind cuisines. Most of the cuisine categories in this dataset are popular foods in Japan and other countries. There are 31,395 images of 256 cuisines. Following the train / test split in~\cite{liu2019few}, we use 27,849 images of 224 cuisines as the source set and the rest 3,546 images of another 32 cuisines as the target set. In the target set, we select 512 images as query and the rest 3,034 are used as the database images. There is no overlapping class (food category) between the source and target sets.   

\noindent\textbf{CUB-200-2011} consists of $11,788$ images of $200$ bird species~\cite{WahCUB_200_2011}. Following~\cite{Song2016Deep}, we use the CUB-200-2011 dataset for fine-grained image retrieval. The first 100 classes (5,864 images) are used as source set and we evaluate the model on the rest 100 classes (5,924 images).

\noindent\textbf{Market-1501} is a large-scale public pedestrian retrieval dataset~\cite{zheng2015scalable}. This type of retrieval task is also known as person re-identification (re-ID), which aims at spotting a person of interest across the camera network. For instance, the technique can help quickly find the lost child in a large park, campus, or airport. The author collects images under six different cameras at a university campus. There are 32,668 detected images of 1,501 identities. Following the standard train / test split, we use 12,936 images of 751 identities as the source set and the rest 19,732 images of another 750 identities as the target set. There is no overlap class (identity) between the source and target sets.   

\noindent\textbf{Oxford5k \& Paris6k} are two widely-used landmark retrieval datasets. Oxford5k contains 5,062 images of 11 particular Oxford buildings~\cite{philbin2007object}, and Paris6k contains 6,412 images of 12 particular Paris landmarks~\cite{philbin2008lost}, respectively. Both datasets are only used as the target set for evaluation. Following the practice in~\cite{radenovic2018fine}, we deploy the non-overlapping building images collected on Flickr as source set to train the model. The source set contains 133k images.

\noindent\textbf{Cifar-10} is a widely-used image recognition dataset, containing 60,000 images of $10$ classes~\cite{Krizhevsky2009Learning}. 
There are 50,000 training images and 10,000 test images. We conduct the image recognition evaluation on this dataset.

\begin{SCfigure*}[][h]
  {\includegraphics[width=1.5\linewidth]{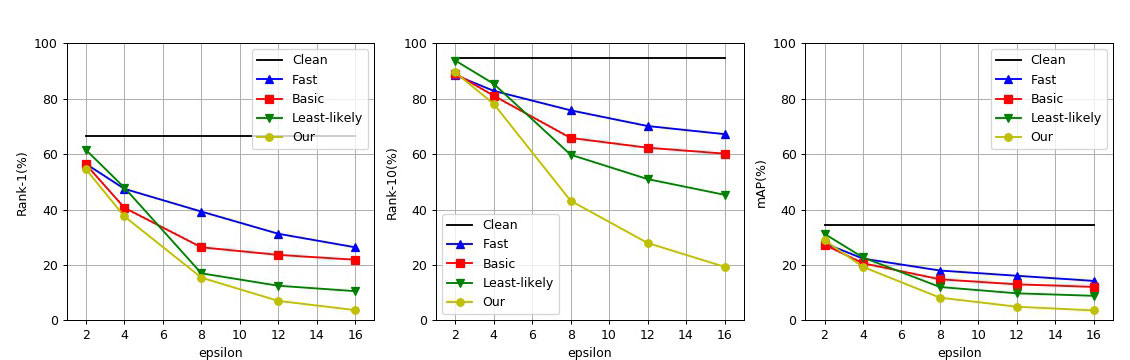}}
   \caption{Recall@1 ($\%$), Recall@10 ($\%$) and mAP ($\%$) of the victim model on Food-256 under the attack by different methods and different perturbation rates $\epsilon$. ``Clean'' denotes the result obtained by using the original query without any attack. The victim model using clean queries arrives at Recall@1 = $66.41\%$, Recall@10 = $94.53\%$ and mAP = $34.56\%$.
   } 
\label{market}
\end{SCfigure*}

\begin{SCfigure*}[][h]
  {\includegraphics[width=1.5\linewidth]{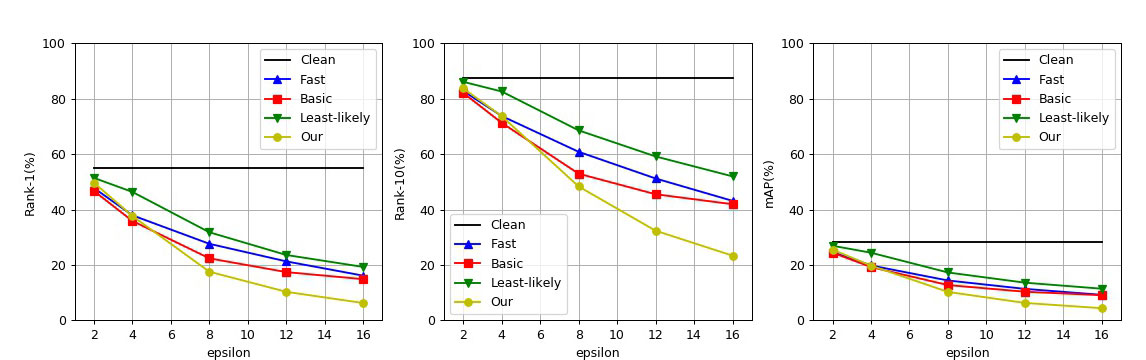}}
   \caption{Recall@1 ($\%$), Recall@10 ($\%$) and mAP ($\%$) of the victim model on CUB-200-2011 under the attack by different methods and different perturbation rates $\epsilon$. ``Clean'' denotes the result by inputting the original query without any attack.  The victim model using clean queries arrives at Recall@1 = $54.86\%$, Recall@10 = $87.51\%$ and mAP = $28.29\%$.
   } 
\label{fig:CUB}
\end{SCfigure*}

\begin{SCfigure*}[][h]
  {\includegraphics[width=1.5\linewidth]{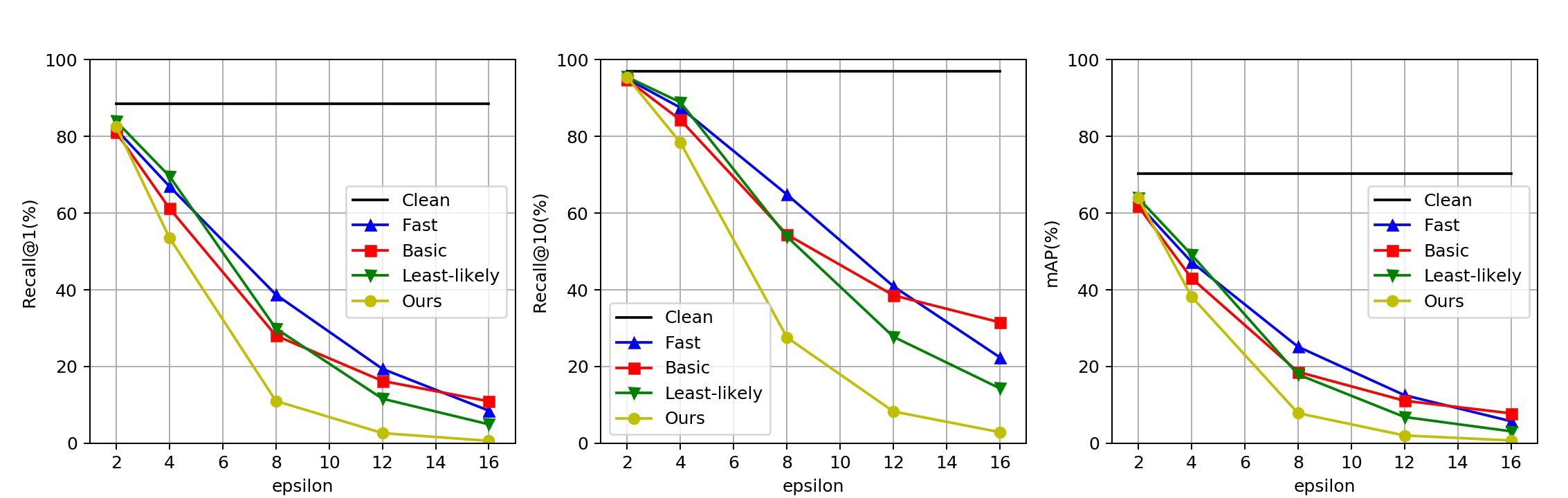}}
   \caption{Recall@1 ($\%$), Recall@10 ($\%$) and mAP ($\%$) of the victim model on Market under the attack by different methods and different perturbation rates $\epsilon$. ``Clean'' denotes the result by inputting the original query without any attack. The victim model using clean queries arrives at Recall@1 = $88.56\%$, Recall@10 = $97.03\%$ and mAP = $70.28\%$.
   } 
\label{fig:market}
\end{SCfigure*}

\noindent\textbf{Evaluation Metric.}
With the limited image perturbation, we compare the methods by the drop of the accuracy. The lower accuracy indicates that the adversarial examples make more true matches receive low ranks. For image retrieval, we use two evaluation metrics, \emph{i.e.,}  Recall@K and mean average precision (mAP). \textbf{Recall@K} is the probability that the right match appears in the top-K of the ranking list. Given a ranking list, the average precision (AP) calculates the space under the recall-precision curve. \textbf{mAP} is the mean of the average precision of all queries. 
For image recognition, we use Top-1 and Top-5 accuracy. \textbf{Top-K} is the probability that the right class appears in the top-K predicted classes.

\noindent\textbf{Implementation Details of the Victim Model.}
For the classification-based retrieval victim model, we follow the common practice in~\cite{hermans2017defense,chen2017person} to fine-tune the ResNet-50~\cite{he2016deep} by class prediction on Food-256, CUB-200-2011 and Market-1501. 
During training, the cuisine images in Food-256 are resized to $256\times256$, while the pedestrian images of Market-1501 is resized to $256\times128$ following the previous practices~\cite{sun2017beyond,zhong2020learning}. The images in CUB-200-2011 are first resized with its shorter side to $256$, and we then apply a $256\times 256$ random crop to the images. 
The learning rate is $0.01$ for the first 40 epochs and decays to $0.001$ for the last 20 epochs. 
For the ranking-based retrieval victim model, we follow the setting in~\cite{radenovic2018fine} to train the ResNet-101~\cite{he2016deep} on the collected building dataset~\cite{radenovic2018fine} with contrastive loss. 
For image recognition, our implementation employs ResNet with $20$ layers for the Cifar-10 dataset~\cite{he2016deep}. The size of the input image is $32\times32$.
The training policy follows the practice in~\cite{he2016deep}. The learning rate starts from 0.1 and is divided by 10 after the 150th and 225th epoch. We stop training after 300 epochs.

\begin{table*}[tbp]
\caption{ \textbf{Attack the classification-based retrieval model.} We mainly compare the proposed method with other attack methods on three datasets, \ie Food-256, CUB-200-2011 and Market-1501. Here we show the results in $\%$ (Lower is better). The perturbation rate is fixed to $\epsilon = 16$. We compare the three classification attack methods, \emph{i.e.,} Fast \protect\cite{goodfellow2014explaining}, Basic \protect\cite{kurakin2016adversarial}, Least-likely \protect\cite{kurakin2016adversarial}. }
\begin{center}
{
\label{table:STA}
\setlength{\tabcolsep}{15pt}
\begin{tabular}{l|cc|cc|cc}
\shline
\multirow{2}{*}{Methods} & \multicolumn{2}{c|}{Food-256} & \multicolumn{2}{c|}{CUB-200-2011} & \multicolumn{2}{c}{Market-1501}\\
& Recall@1 & mAP & Recall@1 & mAP & Recall@1 & mAP \\
\hline
Victim & 66.41 & 34.56 & 54.86  & 28.29 & 88.56 & 70.28 \\
\hline
Fast~\cite{goodfellow2014explaining} & 26.37 & 14.25 & 8.74 & 4.88 & 8.49 & 5.74 \\
Basic~\cite{kurakin2016adversarial} & 21.88 & 12.09 & 8.88 &  5.57 & 10.87 & 7.77  \\
Least-likely~\cite{kurakin2016adversarial} & 10.55 & 8.87 & 9.60 & 4.78 & 4.87 & 3.09  \\
\hline
ODFA & \textbf{3.71} & \textbf{3.59} & \textbf{1.81} & \textbf{1.72} & \textbf{0.68} & \textbf{0.72} \\
\shline
\end{tabular}}
\end{center}
\end{table*}

\begin{table}[tbp]
\small
\caption{ \textbf{Attack the ranking-based retrieval model.} We mainly evaluate the attack performance on two datasets, \ie Oxford5k and Paris6k, with and without multiple-scale (MS) evaluation. Here we show the results in $\%$ (Lower is better). The perturbation rate is fixed to $\epsilon = 16$.}
\begin{center}
{
\label{table:Ox}
\setlength{\tabcolsep}{7pt}
\begin{tabular}{l|cc|cc}
\shline
\multirow{2}{*}{Methods} & \multicolumn{2}{c|}{Oxford5k} &\multicolumn{2}{c}{Paris6k} \\
 & Recall@1 & mAP & Recall@1 & mAP \\
\hline
Victim & 100.00 & 86.24 & 100.00 & 90.66  \\
\hline
ODFA & 0.00 & 0.77 & 3.69 & 2.86 \\
\shline
\hline
Victim (MS) & 100.00 & 88.17 & 100.00 & 92.52  \\
\hline
ODFA & 92.73 & 73.80 & 98.18 & 87.98 \\
ODFA-MS & 1.82 & 2.24 & 3.64 & 4.78 \\
\shline
\end{tabular}}
\end{center}
\end{table}

\begin{figure}[t]
\begin{center}
   \includegraphics[width=1\linewidth]{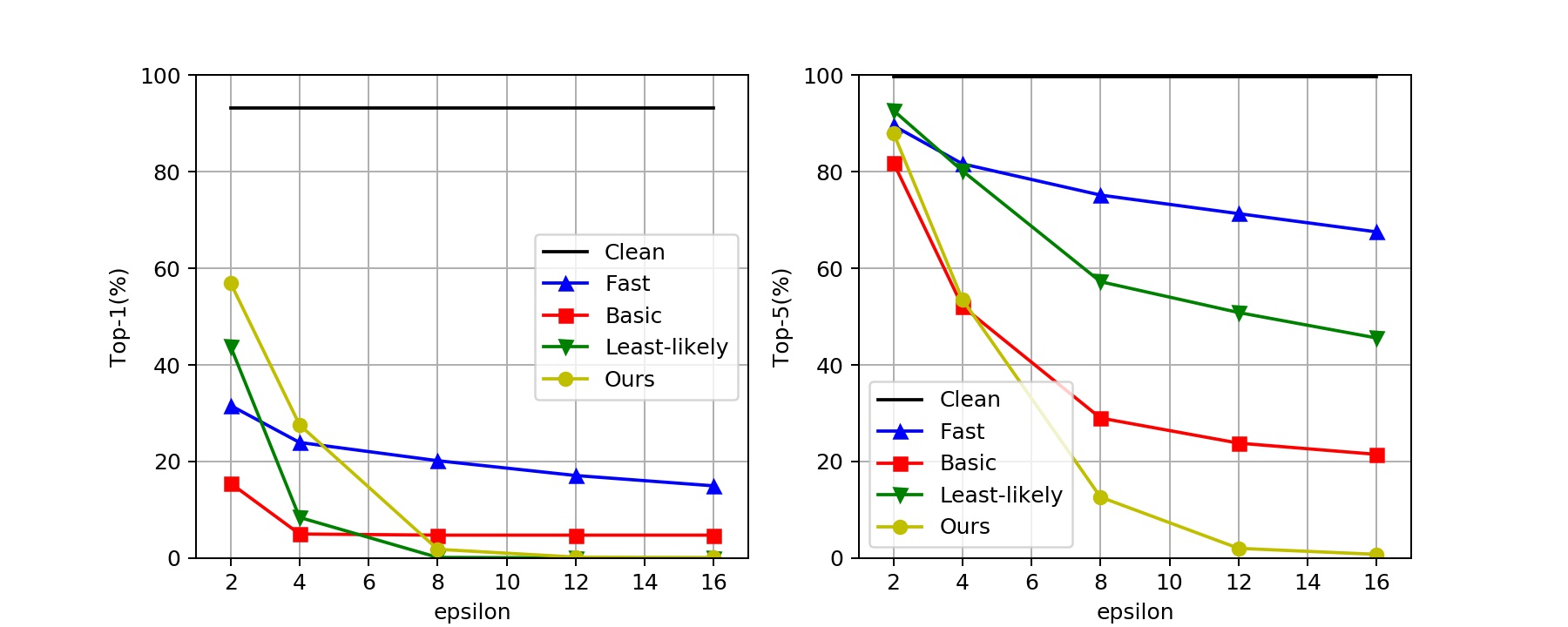}
\end{center}
   \caption{Top-1 ($\%$) and Top-5 ($\%$) accuracy of the victim model on Cifar-10 under the attack by different methods and different perturbation rates $\epsilon$. ``Clean'' denotes the result by using original query images. 
   }
\label{fig:cifar}
\end{figure}

\begin{table}[tbp]
\small
\caption{ \textbf{Attack the image recognition on Cifar-10.} Here we show the results in $\%$ (Lower is better). The perturbation rate is fixed to $\epsilon = 16$. We compare the three classification attack methods, \emph{i.e.,} Fast \protect\cite{goodfellow2014explaining}, Basic \protect\cite{kurakin2016adversarial}, Least-likely \protect\cite{kurakin2016adversarial}. }
\begin{center}
{
\label{tab:cifar}
\setlength{\tabcolsep}{20pt}
\begin{tabular}{l|cc}
\shline
\multirow{2}{*}{Methods} & \multicolumn{2}{c}{Cifar-10}\\
 & Top-1 & Top-5\\
\hline
Victim  & 93.14 & 99.76 \\
\hline
Fast  & 14.95 & 67.55 \\
Basic  & 4.74 & 21.47 \\
Least-likely  & \textbf{0.03} & 45.58 \\
\hline
ODFA  & 0.06 & \textbf{0.76} \\
\shline
\end{tabular}}
\end{center}
\end{table}

\subsection{Effectiveness of ODFA in the Classification-based Retrieval Model}
We first demonstrate the superior attack performance of ODFA to other conventional attack methods. 
The quantitative results, \ie Recall@1, Recall@10 and mAP, on Food-256 using clean and adversarial queries are summarized in Figure~\ref{market}.
The victim model using clean queries arrives at a relatively high performance: Recall@1 = $66.41\% $ and mAP = $34.56\% $. 
As mentioned, the classification attack method changes the semantic prediction, which \emph{implicitly} changes the retrieval features. When the perturbation rate $\epsilon = 8$, the adversarial images generated by the three classification attacks lead to more than $50\%$ rank-1 error. When $\epsilon = 16$, the iterative least-likely class method even yields a Recall@1 = $10.55\%$. 
Nevertheless, these methods are not very effective to move true matches out of the top-10 rank. 
Although Recall@10 continues to decrease with increasing $\epsilon$, the best method, \emph{i.e.,} the iterative least-likely class method, only achieves a Recall@10 of $45.31\%$.
In comparison, the proposed ODFA achieves a lower Recall@1 and Recall@10 when $\epsilon=8$. This can be attributed to the opposite gradient direction attack mechanism. Since the distance between the feature of the adversarial query and that of the original query is much larger, the true matches, which are close to the original query, are thus far from the adversarial query in the feature space. As we increase the perturbation rate $\epsilon$ to $16$, the victim model yields Recall@1 = $3.71\%$, Recall@10=$19.34\%$, mAP = $3.59\%$, which is lower than all the traditional classification attack methods.  



\begin{figure*}[t]
\begin{center}
   \includegraphics[width=1\linewidth]{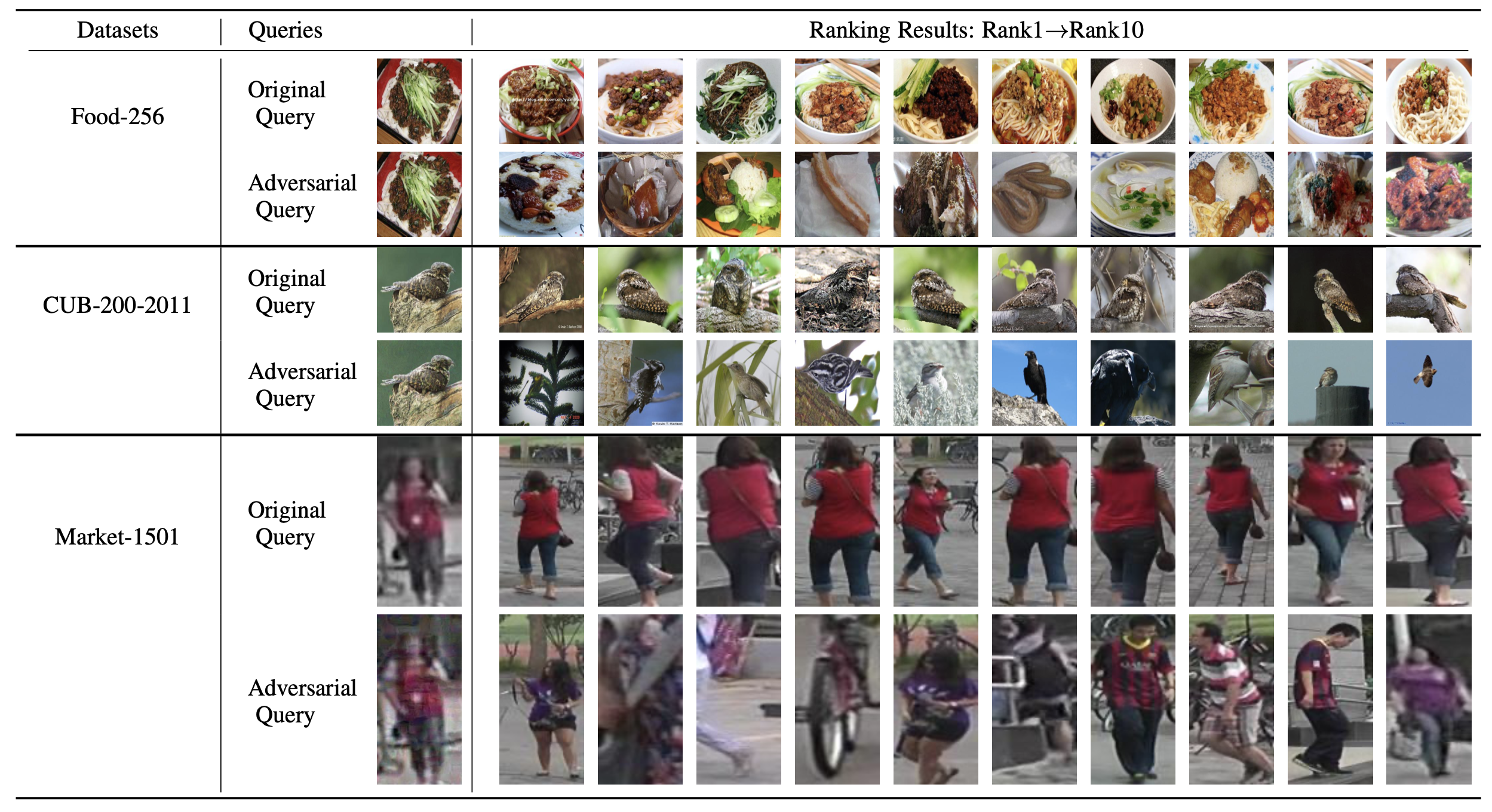}
\end{center}
   \caption{ Ranking results of the original queries and the adversarial queries generated by our method. The proposed approach introduces trivial noise on original queries to fool the retrieval system, while the human is robust to such noise. Three original queries are from Food-256~\cite{kawano14c}, CUB-200-2011~\cite{WahCUB_200_2011} and Market-1501~\cite{zheng2015scalable}, respectively. The corresponding top-10 retrieval results are also provided. The proposed adversarial queries successfully fool the retrieval model to predict irrelevant ranking results. (Best viewed when zoomed in)
   }
\label{fig:0}
\end{figure*}

The experiments on the fine-grained image retrieval dataset, \ie CUB-200-2011, and pedestrian retrieval dataset, \ie Market-1501, indicate similar observation (see Figure~\ref{fig:CUB} and Figure~\ref{fig:market}). First, due to the subtle differences among the fine-grained classes, the baseline victim model does not arrive a relatively high performance: Recall@1 = $54.86\%$, Recall@10 = $87.51\%$ and mAP = $28.29\%$ using clean queries. We, however, still can use the proposed method to make the retrieval accuracy even worse. When $\epsilon = 16$, we arrive at Recall@1 = $1.81\%$, Recall@10 = $ 8.76\%$ and mAP = $1.72\%$. 
Second, compared with the three classification attack methods, our method achieves a larger accuracy drop. Since there are no overlapping bird classes in the source and target sets, the impact of the classification attack is limited. When $\epsilon = 16$, the best classification attack, \emph{i.e.,} fast-gradient sign method arrives at Recall@1=$ 8.74\%$, Recall@10 = $31.79\%$ and mAP = $4.88\%$. This accuracy drop is smaller than the drop of the proposed feature-based method. It shows that the proposed ODFA can more effectively and efficiently fool the target retrieval model with the small, human-imperceptible noise. Similarly, on the Market-1501 dataset, the proposed method successfully fools the victim model of predicting worse ranking predictions. The mAP accuracy drops from $70.28\%$ to $0.72\%$. More quantitative results are shown in Table~\ref{table:STA} . 

\subsection{Effectiveness of ODFA in the Ranking-based Retrieval Model} \label{subsec:ranking}
Ranking-based retrieval models use the distance metric, and do not contain the classification prediction part. The traditional classification attack methods, which depend on category prediction, could not work on this line of retrieval models. We, therefore, only evaluate the proposed method to attack the victim model in Table~\ref{table:Ox}. The victim model using clean queries arrives at a high performance: Recall@1 = $100.00\%$, mAP = $86.24\%$ on Oxford5k and Recall@1 = $100.00\%$, mAP = $90.66\%$ on Paris6k. When $\epsilon = 16$, the proposed method successfully fools the victim model, and the accuracy drops to Recall@1 = $0.00\%$, mAP = $0.77\%$ on Oxford5k and Recall@1 = $3.69\%$, mAP = $2.86\%$ on Paris6k, respectively.

Furthermore, we evaluate ODFA on attacking the multiple-scale inputs. Following the practice in~\cite{radenovic2018fine}, we extract and fuse the features of multiple-scale inputs. The fusion of multiple-scale features leads to a robust representation towards scale variants, and slightly improves the victim retrieval performance. The victim model arrives at Recall@1 = $100.00\%$, mAP = $88.17\%$ on Oxford5k and Recall@1 = $100.00\%$, mAP = $92.52\%$ on Paris6k. We observe that the imperceptible noise generated by ODFA are somehow deprecated after resizing the image, and the attack performance is limited. As shown in Table~\ref{table:Ox}, ODFA works on the original-scale inputs and only achieves a small performance drop. In contrast, the extended ODFA-MS, benefiting of considering the multiple-scale adversarial gradients, successfully fools the victim model. The victim model with multi-scale inputs also drops to a low precision at Recall@1 = $1.82\%$, mAP = $2.24\%$ on Oxford5k and Recall@1 = $3.64\%$, mAP = $4.78\%$ on Paris6k.

\begin{figure*}[t]
\begin{center}
   \includegraphics[width=1\linewidth]{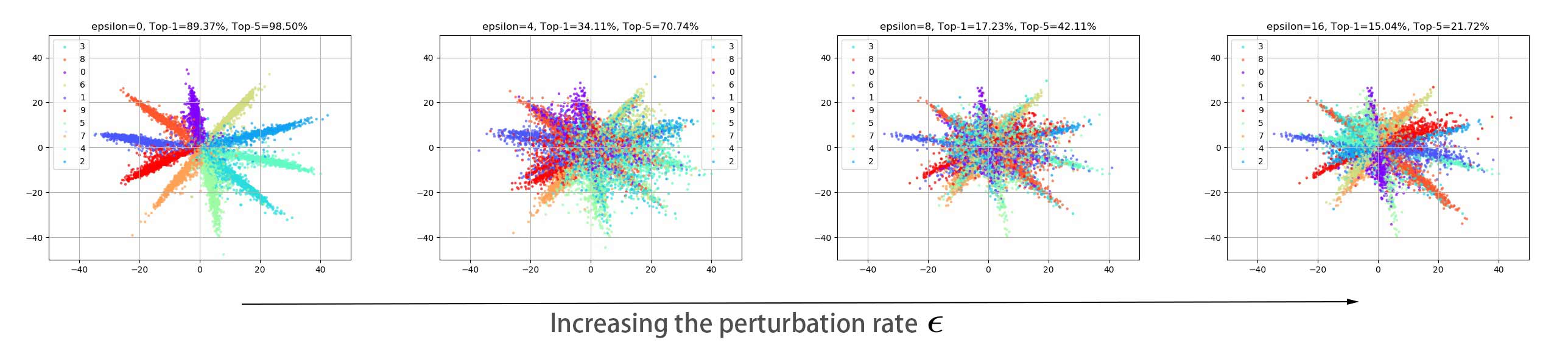}
\end{center}
   \caption{Feature visualization on Cifar-10 (best viewed in color). 
   As the perturbation rate $\epsilon$ increases, the feature gradually moves to the opposite side of the original direction. Top-1 ($\%$) and Top-5 ($\%$) accuracy of the victim model (in the title of each subfigure) also decrease.}
\label{cloud}
\end{figure*}

\subsection{Performance of ODFA in Image Recognition}
We further test ODFA in image recognition. Results are shown in Figure~\ref{fig:cifar}. We observe that our attack does not achieve the largest drop of top-1 accuracy when $\epsilon$ is small. This can be explained by the adversarial target. The iterative least-likely class method aims to make the model mis-classify the adversarial example into the least-likely class. In comparison, our method does not increase the probability of a specific class. Although the confidence score of the correct class decreases, there are no competitors to replace the correct top-1 class which already has a high confidence score. 
Nevertheless, as for top-5 misclassification, the proposed method converges to a lower point than other methods. Since the value of the bias term $b$ for 10 classes is close, we ignore the impact of $b$. When our method converges, the original top-1 prediction $p = Wf$ becomes the lowest probability $p' = -Wf$. So the correct class is moved out of the top-5 classes quickly.
When $\epsilon = 16$, the adversarial images generated by our method compromise the top-5 accuracy from $99.76\%$ to $0.76\%$. The attacked top-1 accuracy $0.06\%$ is also competitive to the result of iterative least-likely class method $0.03\%$. In summary, the proposed ODFA method reports competitive performance and is not evidently superior to the competing methods as the case in image retrieval (see Table~\ref{tab:cifar}). 

\subsection{Effectiveness of ODFA in the Black-box Setting}
As shown in previous works~\cite{szegedy2013intriguing,Papernot2016The,Papernot2016Transferability,papernot2017practical,Liu2017Delving,Moosavidezfooli2017Universal}, adversarial examples have good transferability that can successfully attack other black-box models in the recognition scenario, because the models learn a similar decision boundary in the classification space. In this section, we study the transferability of the adversarial queries in terms of the retrieval scenario. 

For the classification-based retrieval model, we train a stronger victim with \emph{DenseNet-121} \cite{Huang2017Densely} as the black-box model, which arrives at Recall@1 = $89.96\%$ and mAP = $73.39\%$ using ``clean'' images on Market-1501. The adversarial queries are independently generated by the white-box \emph{ResNet-50} ($\epsilon=16$). The experiment shows that adversarial samples generated by \emph{ResNet-50} also compromise the performance of \emph{DenseNet-121}: Recall@1 = $10.24\%$ and mAP = $7.88\%$.


We observe a similar phenomenon on attacking the ranking-based retrieval model. We train the white-box model with \emph{ResNet-101} and use the \emph{ResNet-101} generated adversarial query to attack the black-box model based on \emph{VGG-16}~\cite{simonyan2014very}. The generation process of the adversarial queries are totally independent with the black-box model. The accuracy of the black-box model also drops from $100.00\%$ to $0.00\%$ Recall@1 and $85.24\%$ to $0.79\%$ mAP on Oxford5k. It verifies that the adversarial queries have good transferability and could also be applied to the black-box setting.

\begin{figure*}[t]
\begin{center}
    \includegraphics[width=1\linewidth]{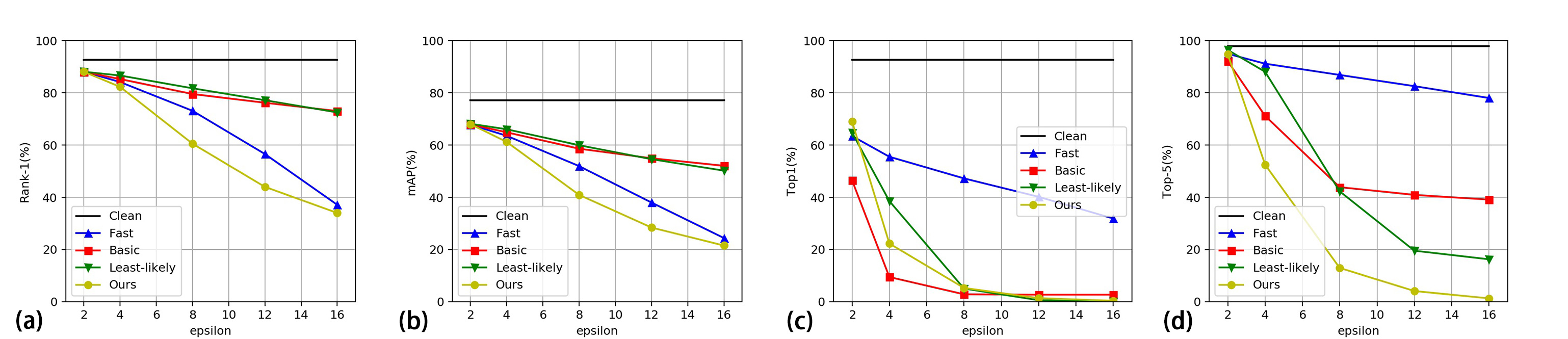}
\end{center}
     \caption{ Performance of attacking state-of-the-art models. (a) and (b): Recall@1 (\%) and mAP(\%) on Market-1501 when attacking the victim model \emph{PCB}~\cite{sun2017beyond}. (c) and (d): Top-1 and Top-5  accuracy (\%) on Cifar-10 when attacking WideResNet-28~\cite{zagoruyko2016wide}.
     }\label{fig:result}
\end{figure*}

\subsection{Further Analysis and Discussions}
\noindent\textbf{Attack against State-of-the-art Methods.}
Furthermore, we evaluate our method on some state-of-the-art models, which achieve competitive accuracy on benchmarks. 
Specifically, for person retrieval (image retrieval), we attack a recent ECCV'18 model called \emph{PCB}~\cite{sun2017beyond}. 
On Market-1501, our re-implementation arrives Recall@1 = $92.70\%$, mAP = $77.14\%$ using clean queries for the victim model. As shown in Figure~\ref{fig:result}~(a,b), Recall@1 and mAP drops to $34.00\%$ and $21.52\%$ respectively by the proposed ODFA. The second best method, Fast-gradient sign method, also arrives a relatively low accuracy $37.11\%$  and $24.40\%$, but is still smaller than the accuracy drop of the proposed method.
For image recognition, we evaluate our method on the prevailing WideResNet-28~\cite{zagoruyko2016wide}. Our re-implementation arrives Top-1 accuracy $96.14\%$ and Top-5 accuracy $99.91\%$ using clean queries, respectively. As shown in Figure~\ref{fig:result}~(c,d), we have consistent observations with the baseline victim models, \emph{i.e.,} competitive top-1 accuracy drop and largest top-5 accuracy drop. Our method arrives Top-1 accuracy of $0.34\%$ and Top-5 accuracy of $1.29\%$. 

\noindent\textbf{Visualization of Retrieval Results.}
We provide one qualitative comparison on the retrieval results with original queries and adversarial queries in Figure~\ref{fig:0}. Since we employ an iterative policy with small steps, the adversarial queries generated by our method are visually close to the original query, which simulates extreme retrieval cases to evaluate the model robustness. In these examples, the ranking results obtained by the original queries are good. However, when using the adversarial queries, the top-10 ranked images are all false matches with a significantly different appearance to the query. The adversarial query successfully makes the victim model predict low ranks to the true matches.

\noindent\textbf{Visualization of Attacked Features.}
Following the visualization trick in~\cite{liu2016large}, we insert an additional 2-dim fully-connected layer into the CNN model to visualize the feature. We train the victim model with an extra fully-connected layer on Cifar-10 and then extract the 2-dim feature of every test image to plot maps. Due to applying the visualization trick (using the 2-dim feature to classify 10 classes), the accuracy of the new victim model is a little bit lower than the baseline result in Table~\ref{tab:cifar}, but still arrives at a relatively high accuracy, Top-1=$89.37\%$, and Top-5=$98.50\%$. It is good enough to verify our intuition in the feature space. As shown in Figure~\ref{cloud}, the points in the same color belong to the same class. We plot four maps with different perturbation rates $\epsilon = 0,4,8,16$ to see the feature movement. $\epsilon = 0$ is the output of the victim model on clean test images. The features gradually move to the opposite side of the original direction, when $\epsilon$ increases. The observation verifies the effectiveness of our objective, \ie moving to the opposite direction. Comparing the figure of $\epsilon = 0$ with the figure of $\epsilon=16$, the feature of most adversarial examples successfully move to the opposite side of the original feature. Due to the change of the immediate features, the classification accuracy, as shown in the title of every subfigure, also gradually drops. The observation validates the mechanism of the proposed method.

\section{Conclusion}
In this paper, we consider the adversarial attack in a new realistic setting, \emph{i.e.,} image retrieval, and propose a new attack method named Opposite-Direction Feature Attack (ODFA) tailored for the retrieval scenario. Different from previous works, the proposed attack method does not depend on the category prediction. Instead, ODFA takes the advantage of the intermediate feature and explicitly considers the feature distance in the representation space. 
On five image retrieval datasets, \emph{i.e.,} Food-256~\cite{kawano14c}, CUB-200-2011~\cite{WahCUB_200_2011}, Market-1501~\cite{zheng2015scalable}, Oxford5k~\cite{philbin2007object} and Paris6k~\cite{philbin2008lost}, we validate the effectiveness of the proposed method on two kinds of retrieval victims, \ie classification-based retrieval model and ranking-based retrieval model. The proposed ODFA leads to a large performance drop in ranking accuracy with human imperceptible perturbation. We also extend the ODFA to adapt the multi-scale evaluation and verify the effectiveness of ODFA on black-box models. Moreover, we visualize the change of the feature direction to further support our intuition in the feature space.
In the future, we will investigate into applying the proposed attack to shallow layers of neural networks and study its effect on other tasks, such as semantic segmentation, action recognition and object detection~\cite{xie2017adversarial,liang2015proposal,jiang2017exploiting,yang2016discriminative}. 

{\small
\bibliographystyle{ieee}
\bibliography{ieee}
}

\begin{IEEEbiography}[{\includegraphics[width=1in,height=1.25in,clip,keepaspectratio]{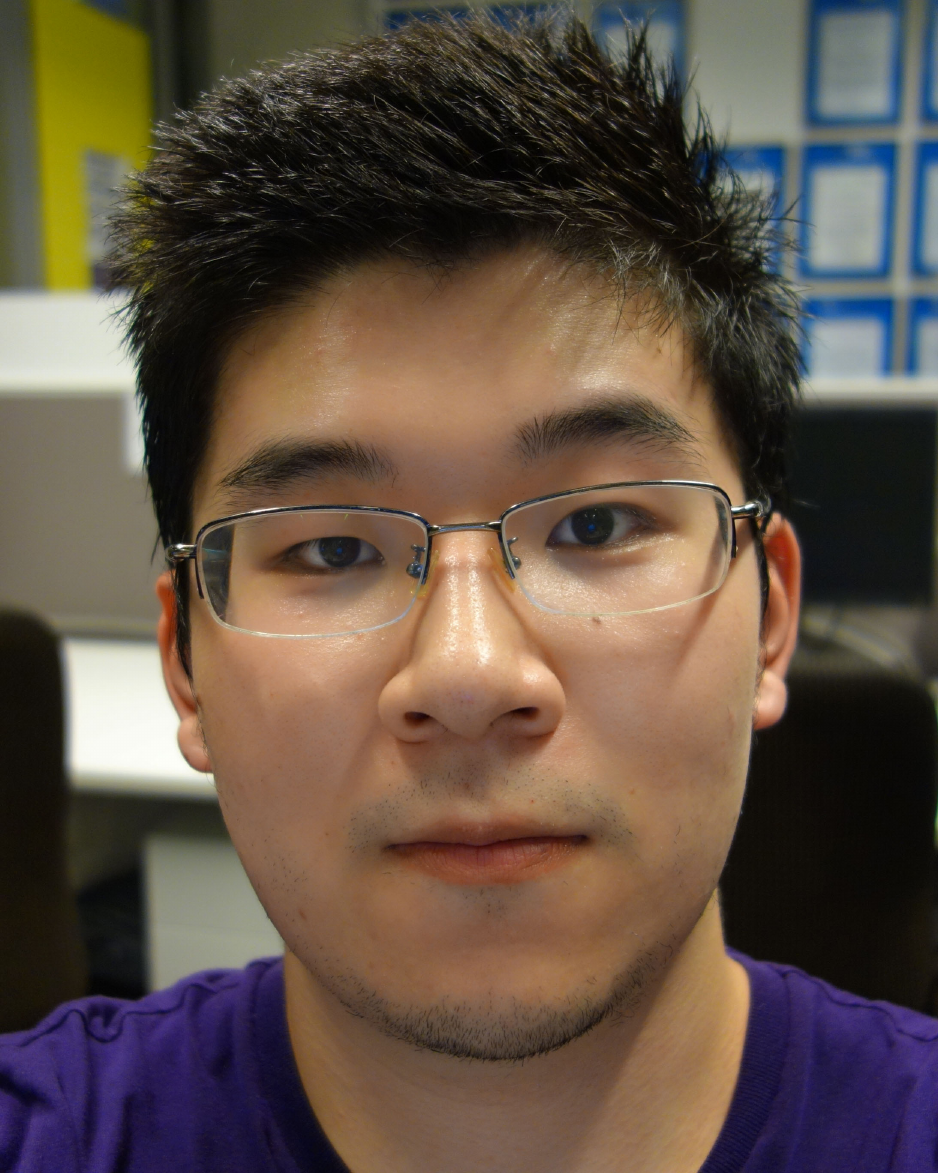}}]{Zhedong Zheng}
received the B.S. degree in computer science from Fudan University, China, in 2016. He is currently a Ph.D. student with the School of Computer Science at University of Technology Sydney, Australia. His research interests include robust learning for image retrieval, generative learning for data augmentation, and unsupervised domain adaptation.
\end{IEEEbiography}
\vfill
\vspace{-.5in}
\begin{IEEEbiography}[{\includegraphics[width=1in,height=1.25in,clip,keepaspectratio]{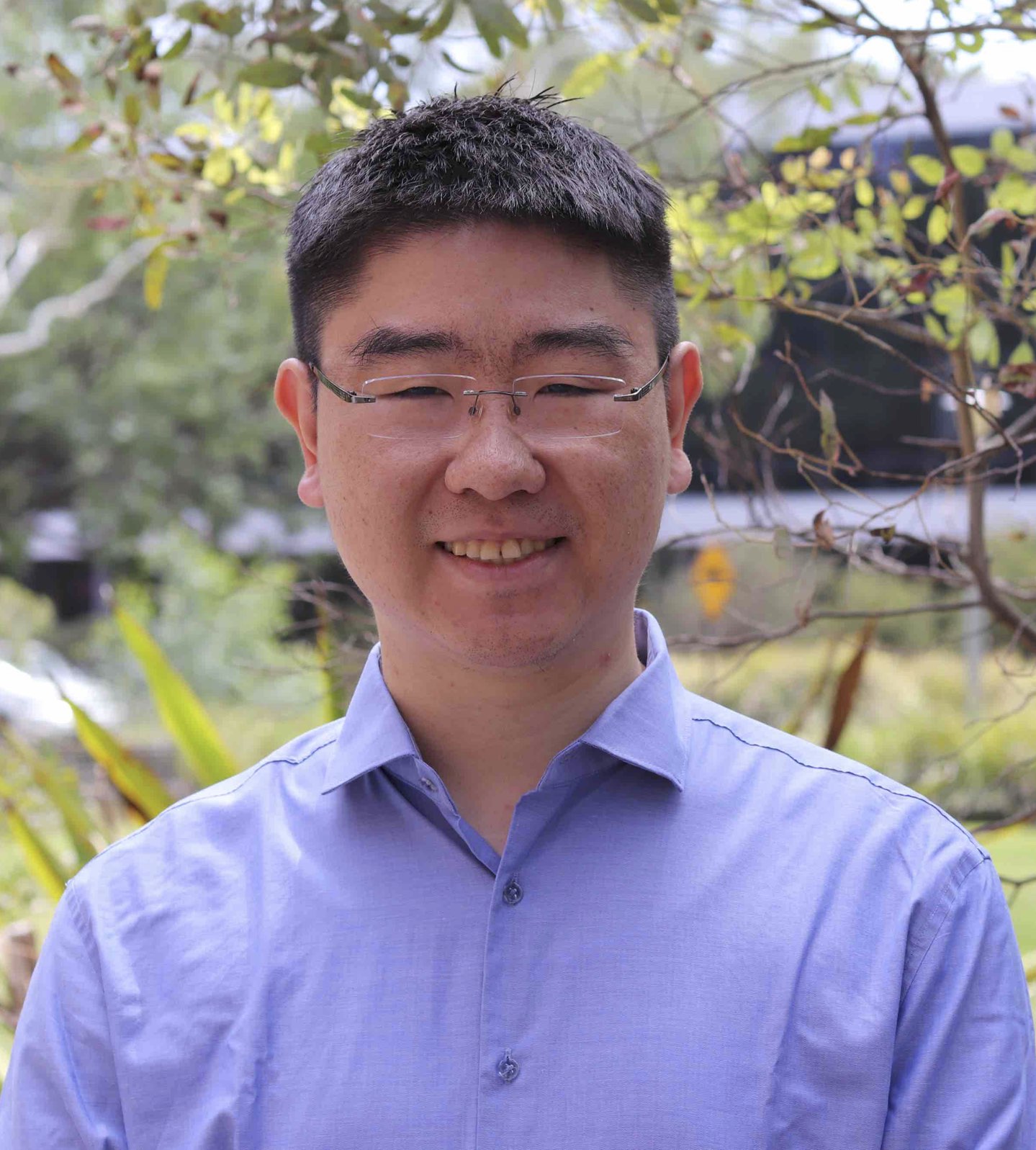}}]{Liang Zheng} is a Lecturer and a Computer Science Futures Fellow in the Research School of Computer Science, Australian National University. He received the Ph.D degree in Electronic Engineering from Tsinghua University, China, in 2015, and the B.E. degree in Life Science from Tsinghua University, China, in 2010. He was a postdoc researcher in the Centre for Artificial Intelligence, University of Technology Sydney, Australia. His research interests include image retrieval, classification.
\end{IEEEbiography}
\vfill
\vfill
\vspace{-.5in}
\begin{IEEEbiography}[{\includegraphics[width=1in,height=1.25in,clip,keepaspectratio]{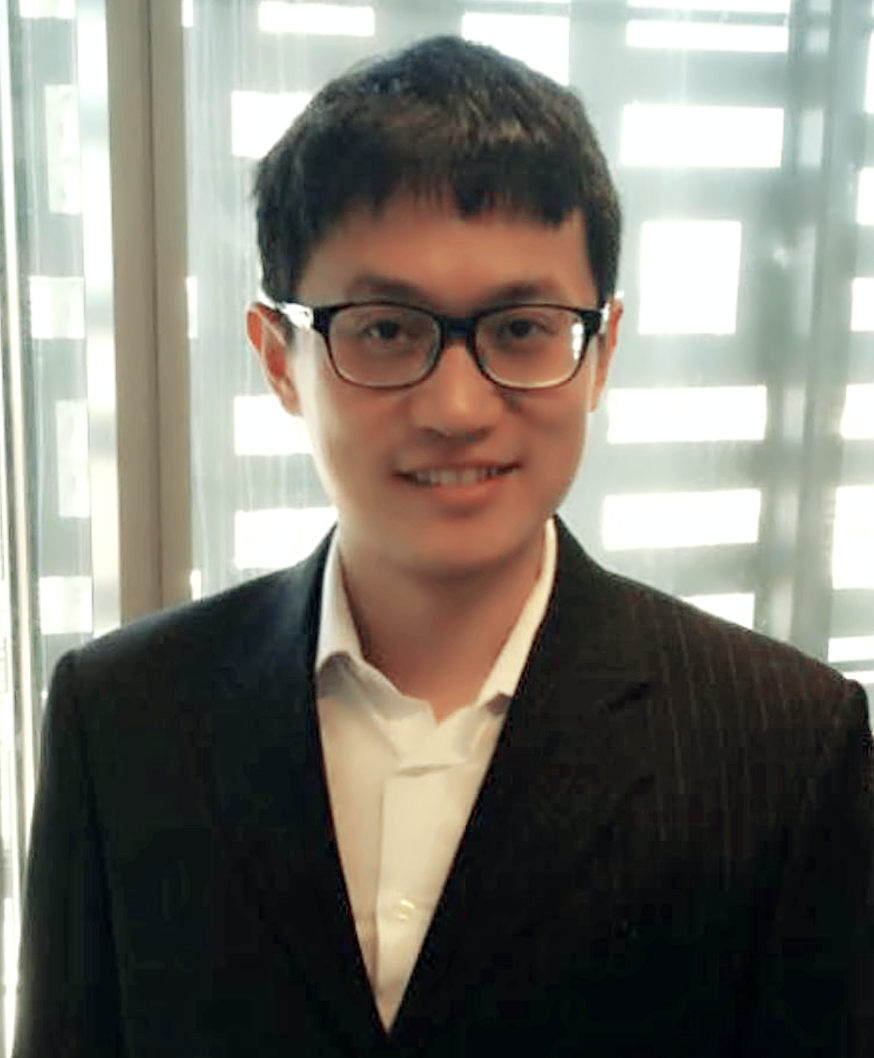}}]{Yi Yang} received the Ph.D. degree in computer
science from Zhejiang University, Hangzhou, China, in 2010. He is currently a professor with University of Technology Sydney, Australia.
He was a Post-Doctoral Research with the School of Computer Science, Carnegie Mellon University, Pittsburgh, PA, USA. His current research interest includes machine learning and its applications to multimedia content analysis and computer vision, such as multimedia analysis and video semantics understanding.
\end{IEEEbiography}

\vfill
\vspace{-.5in}
\begin{IEEEbiography}[{\includegraphics[width=1in,height=1.25in,clip,keepaspectratio]{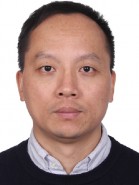}}]{Fei Wu} received his PhD degree in Computer
Science from Zhejiang University in 2002. He
is currently a full professor with the College of
Computer Science and Technology, Zhejiang University. His current research interests include
artificial intelligence, multimedia retrieval and
machine learning. 
\end{IEEEbiography}

\end{document}